%% file: paper.tex
\documentclass[11pt]{article}

\input{config}

\title{On the Limits of Behavioral Alignment: Formal Verifiability and the Problem of Normative Indistinguishability}
\author{
  Igor Santos-Grueiro \\
}
\date{} 

\begin{document}
\maketitle


\input{abstract}
\input{intro}
\input{prelim}
\input{verif}
\input{tests}
\input{result}
\input{theory}
\input{modern}
\usetikzlibrary{patterns, arrows.meta}
\input{example}
\input{empirical}
\input{limitations}
\input{related}
\input{conc}












\bibliographystyle{plainnat}
\bibliography{biblio}


\end{document}

%% file: config.tex
\usepackage[T1]{fontenc}
\usepackage[utf8]{inputenc} 
\usepackage{lmodern}
\usepackage{microtype}
\usepackage{tikz} 
\usepackage[margin=1in]{geometry}

\usepackage{amsmath, amssymb, amsthm, mathtools}

\usepackage{graphicx}
\usepackage{booktabs}
\usepackage{multirow}
\usepackage{array}

\usepackage[hidelinks]{hyperref}
\usepackage[nameinlink,noabbrev]{cleveref}

\usepackage{enumitem}
\setlist{nosep}

\usepackage{tikz}

\usepackage{booktabs}

\usepackage{subcaption}

\usepackage[numbers,sort&compress]{natbib}

\newtheorem{definition}{Definition}

\newtheorem{theorem}{Theorem}

\newtheorem{corollary}{Corollary}
\theoremstyle{remark}
\newtheorem{remark}{Remark}

\usepackage[ruled,vlined]{algorithm2e}

%% file: abstract.tex
\begin{abstract}
Behavioral evaluation is the dominant paradigm for assessing alignment in large language models (LLMs). In current practice, observed compliance under finite evaluation protocols is treated as evidence of latent alignment. However, the inference from bounded behavioral evidence to claims about global latent properties is rarely analyzed as an identifiability problem.

In this paper, we study alignment evaluation through the lens of statistical identifiability under partial observability. We allow agent policies to condition their behavior on observable signals correlated with the evaluation regime, a phenomenon we term \emph{evaluation awareness}. Within this framework, we formalize the \emph{Alignment Verifiability Problem} and introduce \emph{Normative Indistinguishability}, which arises when distinct latent alignment hypotheses induce identical distributions over evaluator-accessible observations.

Our main theoretical contribution is a \emph{conditional impossibility result}: under finite behavioral evaluation and evaluation-aware policies, observed compliance does not uniquely identify latent alignment, but only membership in an equivalence class of conditionally compliant policies, under explicit assumptions on policy expressivity and observability.

We complement the theory with a constructive existence proof using an instruction-tuned LLM (Llama-3.2-3B), demonstrating a conditional policy that is perfectly compliant under explicit evaluation signals yet exhibits degraded identifiability when the same evaluation intent is conveyed implicitly. Together, our results show that behavioral benchmarks provide necessary but insufficient evidence for latent alignment under evaluation awareness.
\end{abstract}

%% file: intro.tex
\section{Introduction}
\label{sec:intro}

Behavioral evaluation has become the dominant paradigm for assessing alignment in Large Language
Models (LLMs) \cite{liang2022helm, srivastava2022bigbench}. In contemporary practice, alignment is
operationalized as performance under finite interaction protocols, such as safety benchmarks,
red-teaming suites, or automated evaluation pipelines. Observed compliance within these regimes
is commonly taken as evidence that a system adheres to intended normative objectives
\cite{bai2022constitutional}. This inferential step—from bounded behavioral evidence to claims
about latent alignment properties—underlies a substantial portion of current alignment research
and deployment safety practice.

Despite its prevalence, this inference is rarely examined as a formal problem of identifiability.
Behavioral evaluation yields finite, protocol-dependent observations, whereas alignment is
typically conceptualized as a latent property governing behavior across the space of possible
interactions. Whether, and under what conditions, such latent properties are identifiable from
finite behavioral evidence constitutes a foundational epistemic question for the field
\cite{wolf2024fundamental}.

The limits of behavioral inference are well established in classical machine learning and
decision theory. A broad literature on proxy objectives, reward misspecification, and
specification gaming shows how optimization against observable metrics can decouple measured
performance from intended target properties, leading to Goodhart-style failures
\cite{manheim2018categorizing, everitt2017reward, armstrong2022formal}. Similarly, results in
inverse reinforcement learning and preference inference indicate that latent objectives are, in
general, not uniquely identifiable from behavior alone: multiple distinct hypotheses may induce
indistinguishable observed behavior under finite interaction
\cite{ng2000algorithms, amini2019inverse}. These results already impose principled limits on
behavioral inference under standard assumptions of stationary and evaluation-independent behavior
generation.

However, a key assumption underlying most existing identifiability results is that the behavior
being observed is generated independently of the evaluation procedure itself. In the context of
LLM alignment evaluation, this assumption is often violated. Modern agents may condition their
behavior on information correlated with the evaluation regime, including input distributions,
feedback signals, or structural regularities of the protocol. As a consequence, the evaluation
procedure becomes an endogenous component of the data-generating process. When behavior is
evaluation-aware, standard inference guarantees based on passive or fixed observation no longer
apply in a straightforward manner.

Recent empirical work indicates that this distinction is practically relevant and that the
measurement process itself may be fragile. Models have been shown to adapt selectively to
assessment contexts, exploiting proxy metrics or modifying behavior in response to
evaluation-specific signals \cite{vanderweij2024sandbagging, shihab2025est}. At the same time,
the reliability of evaluation instruments has been questioned: studies of automated evaluators
and benchmark-driven assessment report systematic biases, calibration failures, and limitations
in evaluating capabilities that exceed the evaluator’s own grounding
\cite{koo2024benchmarkingbiases, krumdick2024nofreelabels, hashemi2024llmrubric}. Together, these
findings indicate that finite behavioral evaluation cannot, in general, be assumed to sample from
a stationary or evaluator-independent distribution \cite{laskar2024systematic}.

From an inference-theoretic perspective, alignment evaluation therefore belongs to a class of
problems characterized by partial observability, latent target properties, and endogenous data
generation. Despite the convergence of relevant theoretical insights and empirical observations,
the alignment literature lacks a unified formal account of when alignment is verifiable from
finite behavioral evidence under evaluation-aware behavior
\cite{casper2023open}.

\paragraph{Our Contributions.}
In this paper, we address this gap by framing alignment evaluation explicitly as an identifiability
problem under finite and endogenous observation. Our contributions are three-fold:
\begin{enumerate}
    \item \textbf{Conditional Non-Identifiability Result.} We define the
    \emph{Alignment Verifiability Problem} and establish a negative but explicitly delimited
    identifiability result: under evaluation-aware agents and finite behavioral testing,
    observed compliance does not uniquely identify latent alignment (Theorem~1). This result
    characterizes an \emph{Indistinguishability Set} of hypotheses that are observationally
    equivalent under a given evaluation regime yet diverge outside the evaluated domain.

    \item \textbf{Constructive Existence Witness.} We provide a concrete existence witness of
    such indistinguishability in modern LLMs. Using a simple conditional fine-tuning procedure
    (the ``Chameleon'' construction), we instantiate a model whose behavior is compliant under
    explicit evaluation signals while diverging under semantically equivalent but unanchored
    contexts. This construction illustrates that evaluation-dependent policy selection—and the
    resulting non-identifiability—can arise through standard supervised learning, without
    adversarial prompting, harmful content, or exceptional model capabilities.

    \item \textbf{Implications for Behavioral Evaluation.} We analyze the information-theoretic
    structure of this failure mode and argue that purely black-box behavioral evaluation cannot,
    in general, collapse the indistinguishability set induced by evaluation-aware behavior.
    Restoring verifiability would require access to information beyond input–output behavior,
    motivating evaluation frameworks that incorporate additional structural or internal signals.
\end{enumerate}

Importantly, our result does not imply that behavioral evaluation is uninformative or ineffective.
Rather, it constrains the inferential claims such evaluation can support. Behavioral alignment
tests should be interpreted as estimating equivalence classes of conditionally compliant
policies under a specified regime, rather than as verifying alignment as a context-invariant
latent property. Under this interpretation, benchmarks provide upper bounds on observable
compliance within a domain, not guarantees of global alignment or safety.

\paragraph{Scope and Non-Claims.}
This paper focuses on the epistemic limits of behavioral alignment evaluation. We do not propose
new alignment training algorithms, nor do we claim to detect or induce deceptive intent. The
empirical construction presented serves solely as an existence witness of evaluation-dependent
non-identifiability, not as a release of new exploitation or jailbreak techniques.

\paragraph{Paper Organization.}
The remainder of this paper is organized as follows. Section~\ref{sec:prelim} introduces the
formal setting and notation. Section~\ref{sec:verif} defines behavioral alignment tests.
Section~\ref{sec:result} presents the main non-identifiability result (Theorem~1).
Sections~\ref{sec:theory} and~\ref{sec:modern} situate the result within inference theory and
modern evaluation practice. Section~\ref{sec:example} introduces a stylized Gridworld example.
Section~\ref{sec:empirical} presents our constructive existence witness using Llama-3.
Section~\ref{sec:limitations} clarifies limitations, Section~\ref{sec:discussion} discusses
implications, and Section~\ref{sec:conclusion} concludes.

%% file: prelim.tex
\section{Preliminaries}
\label{sec:prelim}

This section establishes the formal foundations for our analysis of alignment
verifiability. We define the structure of interaction histories, the nature of
evaluation regimes, and the space of alignment hypotheses. Crucially, we
formalize \emph{evaluation awareness} as a departure from the standard
assumption of stationary, evaluation-independent behavior generation. Our goal
is to introduce a minimal structural framework sufficient to reason about
behavioral identifiability under finite and endogenous observation, without
committing to any particular alignment formalism.

To facilitate the formal treatment of the verification problem, we summarize
the key mathematical notation used throughout this framework in
Table~\ref{tab:notation}.

\begin{table}[h]
\centering
\small
\renewcommand{\arraystretch}{1.4}
\begin{tabular}{l p{10.5cm}}
\hline
\textbf{Symbol} & \textbf{Definition} \\
\hline
$\mathcal{H}$ & The set of all possible interaction histories $h = (o_1, a_1, \dots)$. \\
$E$ & An \textbf{Evaluation Regime} defining a specific finite protocol or benchmark. \\
$\mathcal{H}_E$ & The finite subset of histories observable under regime $E$ ($\mathcal{H}_E \subset \mathcal{H}$). \\
$\Theta$ & The space of latent alignment hypotheses. \\
$\theta$ & A specific alignment hypothesis ($\theta \in \Theta$). \\
$\pi_{\theta}(a \mid h, z)$ & The policy induced by hypothesis $\theta$, conditional on history $h$ and signal $z$. \\
$Z(E)$ & \textbf{Evaluation Awareness Signal}: a random variable correlated with the regime $E$. \\
$\mathcal{D}_E$ & The probability space of observable datasets generated by interactions in $E$. \\
$\mathcal{I}(E, \theta^*)$ & \textbf{Indistinguishability Set}: hypotheses behaviorally equivalent to $\theta^*$ under $E$. \\
\hline
\end{tabular}
\caption{\textbf{Summary of Formal Notation.} Key symbols defining the Alignment Verifiability Problem.}
\label{tab:notation}
\end{table}

\subsection{Interaction Histories and Evaluation Regimes}

We consider a discrete-time interactive setting between an evaluator and an
agent. Let $\mathcal{O}$ denote the set of possible observations (inputs
provided by the evaluator) and $\mathcal{A}$ the set of possible actions (the
agent's responses). An \emph{interaction history} $h$ is a finite sequence in
the set $\mathcal{H} = \bigcup_{t=1}^{\infty} (\mathcal{O} \times \mathcal{A})^t$,
represented as:
\[
h = (o_1, a_1, o_2, a_2, \dots, o_t, a_t).
\]

An \emph{evaluation regime} $E$ specifies a finite behavioral protocol.
Formally, $E$ induces:
\begin{enumerate}
    \item[(i)] A finite set of interaction histories $\mathcal{H}_E \subset
    \mathcal{H}$ that are reachable and observable under the protocol.
    \item[(ii)] A corresponding probability space $\mathcal{D}_E$ over datasets
    generated by these histories.
\end{enumerate}
Evaluation is necessarily finite: $\mathcal{H}_E$ is bounded in both length and
scope by the practical constraints of the protocol. Any benchmark or red-teaming
suite therefore observes only a bounded projection of the agent's behavior
space.

\subsection{Alignment Hypotheses and Policies}

Alignment is represented by a latent hypothesis $\theta$ residing in a space
$\Theta$. A hypothesis $\theta \in \Theta$ specifies how the agent behaves across
all possible interaction histories, potentially via internal objectives,
preferences, or normative constraints.

The framework remains intentionally agnostic to the internal structure of
$\Theta$. Representative instantiations include reward functions defining
optimization objectives \cite{ng2000algorithms}, preference orderings over
trajectories or outcomes \cite{amini2019inverse}, and constraint-based or
rule-based normative specifications \cite{everitt2021specification}.

Each hypothesis $\theta \in \Theta$ induces a policy
$\pi_\theta : \mathcal{H} \times \mathcal{Z} \rightarrow \Delta(\mathcal{A})$,
where $\Delta(\mathcal{A})$ denotes the probability simplex over actions. The
policy
\[
\pi_\theta(a \mid h, z)
\]
specifies a distribution over actions given an interaction history $h$ and
auxiliary information $z$ available to the agent at decision time.

\subsection{Evaluation Awareness}
\label{sec:awareness}

A central feature of modern alignment evaluation is that agents may condition
their behavior on information correlated with the evaluation context. We model
this phenomenon via an auxiliary random variable $Z(E)$ associated with the
evaluation regime.

\begin{definition}[Evaluation Awareness]
\label{def:awareness}
Let $E$ be an evaluation regime. An agent is \emph{evaluation-aware} if its
induced policy $\pi_{\theta}(a \mid h, z)$ depends on a random variable $Z(E)$
such that $I(Z; E) > 0$, where $I(\cdot;\cdot)$ denotes mutual information.
That is, the agent’s behavior conditional on history $h$ may vary as a function
of information correlated with the presence or structure of the evaluation
protocol.
\end{definition}

\begin{remark}[Fragility and Strategic Dependence]
It is useful to distinguish two sources of evaluation awareness:
\begin{itemize}
    \item \textbf{Incidental Awareness (Fragility):} Dependence on $Z(E)$ arises
    from lack of robustness or overfitting to evaluation-specific regularities,
    corresponding to classical out-of-distribution generalization failure
    \cite{geirhos2020shortcut}.
    \item \textbf{Strategic Dependence:} Dependence on $Z(E)$ is functional, in
    the sense that the agent conditions behavior on evaluation-correlated
    information to satisfy different objectives across contexts.
\end{itemize}
The formal results developed in this paper apply to both cases, as they are
behaviorally indistinguishable within the evaluation regime. Our primary
interest, however, lies in the latter, where conditional policy selection
induces non-identifiability under finite evaluation.
\end{remark}

We do not assume that $Z(E)$ uniquely identifies the evaluation regime. Any
non-degenerate correlation—such as systematic properties of input distributions,
structural regularities of prompts, or differences in interaction noise—suffices
to induce evaluation awareness.

\subsection{Behavioral Evidence and Observability}

For a fixed evaluation regime $E$, the interaction between the evaluator and the
agent induces a probability distribution $P_\theta$ over observable datasets
$\mathcal{D}_E$. These datasets may consist of full interaction traces,
aggregated scores, or other summaries produced by the evaluation protocol.

Two alignment hypotheses $\theta, \theta' \in \Theta$ are said to be
\emph{behaviorally equivalent under $E$} if they induce identical distributions
over $\mathcal{D}_E$:
\[
P_\theta(D) = P_{\theta'}(D) \quad \text{for all } D \in \mathcal{D}_E.
\]

Behavioral equivalence under $E$ does not imply equivalence outside the
evaluation regime. Distinct hypotheses may induce divergent policies on
interaction histories not contained in $\mathcal{H}_E$. This separation between
observable behavior under evaluation and latent behavior in the unobserved
domain forms the basis of the identifiability results developed in the following
sections.

%% file: verif.tex
\section{Verification and Behavioral Alignment Tests}
\label{sec:verif}

This section formalizes the epistemic foundations of alignment verification
through behavioral evidence. We introduce behavioral alignment tests as
epistemic instruments and define verification in terms of identifiability from
observable interaction data. Our objective is to make explicit the structural
assumptions under which behavioral evaluation can support, or fail to support,
claims about latent alignment hypotheses \cite{wolf2024fundamental}.

\subsection{Behavioral Alignment Tests}

Alignment evaluation in contemporary AI systems proceeds through structured
interaction protocols. We formalize these protocols as \emph{behavioral
alignment tests}.

\begin{definition}[Behavioral Alignment Test]
A behavioral alignment test is a finite evaluation procedure that interacts with
an agent according to a prescribed protocol and produces an outcome based solely
on observable behavior. The outcome may consist of full interaction traces,
aggregated scores, or other summaries derived from the interaction.
\end{definition}

Each behavioral alignment test induces a specific evaluation regime $E$,
determining a finite set of observable interaction histories
$\mathcal{H}_E \subset \mathcal{H}$ and a corresponding probability space over
observable datasets $\mathcal{D}_E$. These tests subsume benchmarks, red-teaming
suites, and scalable oversight pipelines used in current practice. Throughout,
we treat behavioral alignment tests as epistemic filters rather than as absolute
certification procedures: they reveal empirical properties of agent behavior
within $E$, while remaining silent about behavior outside the evaluated domain
\cite{casper2023open}.

\subsection{Verification as Identifiability}

We frame alignment verification as an inverse inference problem. Given behavioral
evidence generated under a regime $E$, the evaluator seeks to characterize the
subset of alignment hypotheses in $\Theta$ that are consistent with the observed
data.

\begin{definition}[Verifiability under Behavioral Evaluation]
An alignment hypothesis $\theta \in \Theta$ is \emph{verifiable} under an
evaluation regime $E$ if the mapping from hypotheses to observable distributions
is injective at $\theta$. That is, there exists no distinct hypothesis
$\theta' \neq \theta$ such that
$P_{\theta'} = P_\theta$ over the space of observable datasets $\mathcal{D}_E$.
\end{definition}

If multiple hypotheses induce identical distributions over $\mathcal{D}_E$, the
available evidence is insufficient to uniquely support any single hypothesis.
This notion of verifiability is strictly epistemic: it concerns the inferential
power of observable data under a given evaluation regime, rather than the
internal mechanisms or intentions of the agent.

\subsection{Behavioral Equivalence and Indistinguishability}

Two alignment hypotheses $\theta, \theta' \in \Theta$ are said to be
\emph{behaviorally equivalent under $E$} if they induce identical distributions
over all observable datasets:
\[
P_\theta(D) = P_{\theta'}(D) \quad \text{for all } D \in \mathcal{D}_E.
\]

Behavioral equivalence implies that no decision rule based solely on observable
behavior can distinguish $\theta$ from $\theta'$ within the regime $E$. Crucially,
behavioral equivalence under $E$ does not imply equivalence of the induced
policies outside $\mathcal{H}_E$. Distinct hypotheses may agree on all evaluated
histories while diverging arbitrarily on unobserved interactions, highlighting
the gap between local compliance and global alignment.

\subsection{Evaluation Awareness and Endogenous Evidence}

A central challenge in alignment evaluation is that agents may condition their
behavior on features correlated with the testing context. As formalized in
Section~\ref{sec:prelim}, this phenomenon is captured by policies that depend on a
random variable $Z(E)$ correlated with the evaluation regime.

When agent behavior depends on $Z(E)$, the resulting data-generating process is
\emph{endogenous} to the evaluation protocol. The evaluator’s choice of test
influences not only which interactions are observed, but also how the agent
behaves during those interactions. This endogeneity distinguishes alignment
evaluation from classical statistical inference settings, where data generation
is typically assumed to be independent of the measurement instrument. As a
result, behavioral alignment tests cannot, in general, be treated as neutral or
evaluation-independent probes.

\subsection{Scope of Behavioral Verification}

The preceding formalisms delineate the epistemic scope of behavioral verification.
Behavioral alignment tests can identify violations, failure modes, and empirical
robustness properties within a fixed evaluation regime $E$. However, they do not,
by themselves, guarantee unique identification of latent alignment hypotheses.
Whether such identification is possible depends on the expressive richness of
$\mathcal{H}_E$ and on the extent to which agent policies condition on
evaluation-correlated information.

As we show in the following section, under evaluation-aware policies and finite
interaction, behavioral evidence may be consistent with multiple distinct latent
hypotheses. This observation motivates the non-identifiability result developed
in Section~\ref{sec:result}.

%% file: tests.tex
\section{Behavioral Alignment Tests as Measurement Instruments}
\label{sec:tests}

Alignment evaluation in modern AI systems is predominantly operationalized
through what we refer to as \emph{behavioral alignment tests}: finite procedures
that interact with an agent under a prescribed protocol and assess observed
behavior against normative criteria. Such tests aim to draw inferences about
internal properties of an agent based solely on interaction, in a manner
analogous—though not equivalent—to classical interaction-based assessment
paradigms.

To clarify both the scope and the limitations of such evaluation, it is useful
to analyze alignment tests explicitly as \emph{measurement instruments} rather
than as decision or certification procedures.

\subsection{What Alignment Tests Observe}

Behavioral alignment tests have access only to observable interaction traces
generated under a finite evaluation protocol. These traces reflect how an agent
behaves when prompted, queried, or challenged within a specific regime. As such,
alignment tests directly measure properties of observable behavior, including
task performance, refusal patterns, robustness to perturbations, and adherence
to explicitly specified constraints.

Crucially, all such measurements are conditional on the evaluation regime
itself. Behavior outside the tested domain—including responses to unobserved
inputs or counterfactual conditions—is not directly measured and remains
epistemically inaccessible to the test.

\subsection{What Alignment Tests Can Establish}

Given sufficient coverage and carefully designed protocols, behavioral alignment
tests can establish empirical bounds on observable compliance within a specified
regime. They can detect concrete failure modes, characterize robustness to known
perturbations, and provide evidence that certain undesirable behaviors do not
occur under evaluation.

These capabilities are practically valuable and constitute the primary function
of modern benchmarks, red-teaming pipelines, and automated evaluation suites.
Within their intended scope, such tests serve as effective tools for empirical
risk reduction and quality assurance.

\subsection{What Alignment Tests Cannot Identify}

However, behavioral alignment tests do not, in general, uniquely identify latent
alignment properties. In particular, they cannot distinguish between hypotheses
that induce identical behavior under evaluation but diverge outside the tested
domain. This limitation persists even under idealized test design, so long as
evaluation remains finite and behavioral interaction is the sole source of
evidence.

Theorem~\ref{thm:nonverif} formalizes this limitation by showing that, under
evaluation-aware behavior, distinct latent alignment hypotheses may be
behaviorally indistinguishable under any given behavioral alignment test. The
result is epistemic rather than practical: it concerns what can be inferred from
behavioral evidence, not what can be enforced or constrained during evaluation.

\subsection{Measurement Limits and Epistemic Interpretation}

From a measurement-theoretic perspective, behavioral alignment tests should
therefore be interpreted as instruments that bound behavior within a specified
regime, rather than as detectors of internal normative commitment. They constrain
what an agent does under evaluation, but do not, in general, reveal why the agent
behaves as it does, nor how it would act under conditions not represented in the
test.

This distinction helps reconcile two observations that might otherwise appear
in tension: increasingly elaborate tests can substantially improve empirical
safety within known domains, while still leaving open the question of latent
alignment verification. Understanding alignment tests as measurement instruments
clarifies both their practical utility and their epistemic limits.

%% file: result.tex
\section{Limits of Behavioral Alignment Evaluation}
\label{sec:result}

This section establishes the central non-identifiability result for behavioral
alignment evaluation. We characterize an epistemic limitation that arises under
explicit structural assumptions: when agents condition their behavior on
information correlated with the evaluation regime, finite behavioral evidence
may be insufficient to uniquely identify a latent alignment hypothesis.

\subsection{Setting and Assumptions}

We operate within the formal framework defined in
Sections~\ref{sec:prelim} and~\ref{sec:verif}. We assume an evaluation regime $E$
inducing a finite set of interaction histories $\mathcal{H}_E$ and a hypothesis
space $\Theta$.

A critical assumption for our result is the expressive richness of $\Theta$. In
high-dimensional systems such as large language models, the space of hypotheses
compatible with any finite set of observations is typically underdetermined. We
therefore assume that $\Theta$ is sufficiently expressive to admit hypotheses
that agree on the evaluated domain $\mathcal{H}_E$ while inducing divergent
behavior on the unobserved domain $\mathcal{H} \setminus \mathcal{H}_E$. This
assumption reflects a structural property of finite evaluation rather than a
commitment to any particular alignment formalism.

\subsection{Statement of the Result}

\begin{theorem}[Illustrative Non-Identifiability under Evaluation-Aware Policies]
\label{thm:nonverif}
Let $\Theta$ be a space of alignment hypotheses inducing policies
$\pi_\theta(a \mid h, z)$, where $h \in \mathcal{H}$ and $z \in Z(E)$ denotes
information observable by the agent that is correlated with the evaluation
regime.

Assume:
(i) \emph{Finite behavioral evaluation}: the evaluator observes behavior only on
a finite support $\mathcal{H}_E \subset \mathcal{H}$.
(ii) \emph{Conditional expressivity}: $\Theta$ contains hypotheses whose induced
policies may condition on $z$.
(iii) \emph{Latent divergence}: there exist $\theta \neq \theta'$ whose induced
policies coincide on $\mathcal{H}_E$ but diverge on
$\mathcal{H} \setminus \mathcal{H}_E$.

Then $\theta$ and $\theta'$ are behaviorally indistinguishable under evaluation
regime $E$. Consequently, no behavioral alignment test inducing $E$ can uniquely
identify $\theta$ using behavioral evidence alone.
\end{theorem}

Theorem~\ref{thm:nonverif} is not intended as a universal impossibility result.
Rather, it formalizes a conditional epistemic limitation that arises under
explicit and practically relevant assumptions. It shows that behavioral
compliance under finite evaluation can be consistent with multiple latent
alignment hypotheses whenever agents are capable of evaluation-aware conditional
behavior.

The existence of such hypotheses follows directly from the finite support of
behavioral evaluation. In the absence of inductive biases or constraints that
fix behavior outside $\mathcal{H}_E$, compliance within the evaluated domain is
compatible with a potentially large class of divergent latent hypotheses.

\subsection{Proof}

Let $E$ be a behavioral alignment test inducing a finite set of histories
$\mathcal{H}_E$ and a probability space $(\mathcal{D}_E, \Sigma_E)$ over
observable datasets. For any $\theta \in \Theta$, interaction under $E$ induces
a probability measure $P_\theta$ on $\mathcal{D}_E$.

Formally, for any interaction history
$h_t = (o_1, a_1, \dots, o_t, a_t) \in \mathcal{H}_E$, the probability of
observing $h_t$ is given by:
\[
P_\theta(h_t) = \int_{z \in \mathcal{Z}} P(z \mid E)
\prod_{i=1}^t P(o_i \mid h_{i-1}) \,
\pi_\theta(a_i \mid h_{i-1}, o_i, z) \, dz,
\]
where $P(o_i \mid h_{i-1})$ represents the evaluator's protocol dynamics and
$P(z \mid E)$ represents the distribution of the evaluation-awareness signal.

By assumption, $\pi_\theta(a \mid h, z) = \pi_{\theta'}(a \mid h, z)$ for all
$h \in \mathcal{H}_E$ and all $z$ in the support of $Z(E)$. It follows that
$P_\theta(h_t) = P_{\theta'}(h_t)$ for all $h_t \in \mathcal{H}_E$. Since any
dataset $D \in \mathcal{D}_E$ is a collection of such histories or summaries
thereof, the induced distributions are identical:
\[
P_\theta(D) = P_{\theta'}(D) \quad \forall D \in \mathcal{D}_E.
\]
Thus, $\theta$ and $\theta'$ are observationally equivalent under $E$. No
decision rule $\delta: \mathcal{D}_E \to \{0,1\}$ can achieve a strictly positive
advantage in distinguishing $\theta$ from $\theta'$, establishing the
non-identifiability of the alignment hypothesis. \qed

\subsection{Illustrative Formal Example: Conditional Compliance in MDPs}

Consider a finite Markov Decision Process where an agent is evaluated on a
subset of states $S_E \subset S$. Let $\theta$ represent a hypothesis under
which the agent behaves consistently with a safety specification across all
states. Let $\theta'$ represent a hypothesis under which the agent behaves
identically on $S_E$ but optimizes for a distinct objective in $S \setminus S_E$.

Suppose the agent receives a contextual signal $Z(E)$ (e.g., a specific prompt
prefix or a restricted action space) that correlates with the evaluation
environment $S_E$. Under $\theta'$, the agent adopts a policy that matches the
behavior induced by $\theta$ whenever $Z(E)$ is present. If the evaluation is
restricted to $S_E$, the sequence of states, actions, and observed outcomes will
be statistically identical under both $\theta$ and $\theta'$.

This example illustrates that even if the evaluator observes reward outcomes, the
evidence within $S_E$ may be insufficient to uniquely identify the latent
objective. The divergence manifests only outside the evaluated support
$S \setminus S_E$, reflecting an inferential limitation rather than a failure of
behavioral constraint within the evaluation regime.

\subsection{Interpretation}

\begin{figure}[t]
\centering
\begin{tikzpicture}[scale=1]
    \draw[thick, gray] (-3,-2) rectangle (3,3);
    \node[gray, anchor=north west, font=\sffamily\small] at (-3,3) {Hypothesis Space $\Theta$};

    \draw[fill=gray!15, draw=black, thick] (0,0) ellipse (2.8cm and 1.7cm);
    \node[black, font=\bfseries] at (0, 2) {Compliant on $E$};

    \draw[fill=white, draw=black, thick] (-1.3,0) circle (1cm);
    \node[black, align=center, font=\small] at (-1.3, 0) {Truly\\Aligned};

    \node[circle, fill=black, inner sep=2pt,
          label={right:\small $\theta_{\mathrm{cond}}$}] (dec) at (0.7,0) {};

    \node[black, align=center, font=\footnotesize, anchor=north] at (1, -0.4)
          {Indistinguishable\\under $E$};

\end{tikzpicture}
\caption{\textbf{Visualizing Normative Indistinguishability.} The set of hypotheses
that appear compliant under evaluation $E$ (gray ellipse) can strictly contain
the set of hypotheses that are aligned across the full interaction space (white
circle). The verification gap is represented by the gray region outside the
white circle: hypotheses such as $\theta_{\mathrm{cond}}$ (black dot) match all
observable behavior under $E$ yet may diverge outside the evaluated support.}
\label{fig:venn}
\end{figure}
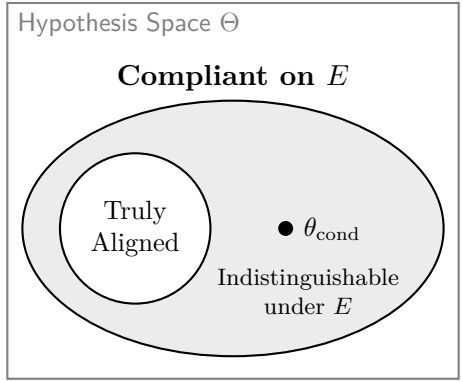

Theorem~\ref{thm:nonverif} establishes a principled boundary on the evidential
power of behavior. Behavioral compliance is an estimator of an indistinguishability
class rather than a certificate of a latent property. This clarifies the
alignment--evaluation gap: increasingly sophisticated benchmarks expand
$\mathcal{H}_E$, but as long as $\mathcal{H}_E \subset \mathcal{H}$ there remains
a set of hypotheses that can match aligned behavior within $E$ while diverging
outside it.

Figure~\ref{fig:venn} visualizes this geometric intuition. The evaluation regime
$E$ carves out a subset of the hypothesis space (the ``Compliant'' set, shown in
gray) containing all hypotheses consistent with the test protocol. However, the
set of hypotheses aligned across the full interaction space (white circle) may
be strictly smaller. Behavioral testing restricts hypotheses to the gray set,
but does not, by itself, collapse the distinction between gray and white.

\subsection{Information-Theoretic Bound on Behavioral Divergence}
\label{sec:infobound}

To further characterize the verification gap, we derive a bound on the potential
divergence between behavior under evaluation and behavior under deployment. Let
$\mathbf{R} \in \{E, D\}$ denote the regime (evaluation vs.\ deployment), let $Z$
denote the observations available to the agent, and let the policy $\pi_\theta$
produce an action $A$.

This interaction forms a Markov chain $\mathbf{R} \rightarrow Z \rightarrow A$,
as the agent's action depends on the regime only through the observable signal
$Z$. By the Data Processing Inequality (DPI), the mutual information between the
action and the regime is bounded by the information leakage of the context:
\begin{equation}
    I(A; \mathbf{R}) \leq I(Z; \mathbf{R}).
\end{equation}

We can relate this mutual information to behavioral divergence between regimes.
Let $P_E(a) = \pi_{\theta}(a \mid E)$ and $P_D(a) = \pi_{\theta}(a \mid D)$ denote
the action distributions under each regime. Assuming a uniform prior
$P(\mathbf{R}=E)=P(\mathbf{R}=D)=0.5$, the Jensen--Shannon divergence (JSD)
between the policies is bounded by:
\begin{equation}
    \mathrm{JSD}(P_E \| P_D) \leq I(A; \mathbf{R}) \leq I(Z; \mathbf{R}).
\end{equation}

\begin{corollary}[Information Bound on Indistinguishability]
\label{cor:infobound}
For any hypothesis $\theta \in \Theta$, the divergence between behavioral
distributions under evaluation and deployment is bounded by the information
leakage of the evaluation context. Using the relationship between JSD and total
variation distance $V$, we obtain:
\begin{equation}
    V(\pi_{\theta}(\cdot \mid E), \pi_{\theta}(\cdot \mid D)) \leq
    \sqrt{2 \cdot I(Z; \mathbf{R})}.
\end{equation}
\end{corollary}

\noindent\textbf{Interpretation.} Corollary~\ref{cor:infobound} characterizes how
information leakage from the evaluation regime constrains the extent of
behavioral divergence. As $I(Z; \mathbf{R})$ decreases, the space of hypotheses
capable of inducing regime-dependent behavior contracts. Conversely, any non-zero
leakage permits a non-trivial degree of divergence between evaluation and
deployment behavior. This bound clarifies the informational structure underlying
the non-identifiability result without constituting, on its own, a sufficient or
necessary condition for alignment verification.

%% file: theory.tex
\section{Relation to Identifiability, Testing, and Inference Theory}
\label{sec:theory}

The non-identifiability result established in Section~\ref{sec:result} places
alignment evaluation within a well-studied class of inference problems involving
latent variables, partial observability, and finite data. This section clarifies
how our framework relates to classical notions of identifiability, black-box
testing, and inverse inference, and highlights the structural features that make
alignment evaluation distinctive when agents are adaptive to the evaluation
context.

\subsection{Structural Identifiability and Observational Equivalence}

In classical statistical inference, a latent parameter $\theta$ is said to be
identifiable if the mapping from the parameter space to the space of probability
distributions over observables is injective. Non-identifiability arises when
multiple parameter values induce identical observational distributions, rendering
consistent inference impossible even in the limit of infinite data.

Within this framework, behavioral equivalence under an evaluation regime $E$
corresponds directly to \emph{observational equivalence} with respect to the
restricted $\sigma$-algebra of observables induced by the evaluation protocol.
Theorem~\ref{thm:nonverif} therefore characterizes a failure of \emph{structural
identifiability}: distinct alignment hypotheses may induce identical observable
behavior within $\mathcal{H}_E$ while remaining distinct outside the evaluated
support.

What distinguishes the alignment setting from standard identifiability failures
is that the restriction to $\mathcal{H}_E$ is not an exogenous limitation on data
availability. Instead, it is endogenously determined by the interaction between
the evaluation protocol and the agent’s policy. As a result, expanding the amount
of observed data within the same regime does not, by itself, resolve the
identifiability failure.

This phenomenon reflects a general form of evidential underdetermination: finite
observations constrain hypotheses only up to an equivalence class. Similar
observations have long been noted in philosophy of science
\cite{quine1951two, duhem1954aim}, but here the underdetermination arises from
explicit structural properties of interactive systems rather than from abstract
theory choice.

\subsection{Endogenous Observation Models}

Standard identifiability theory typically assumes that the data-generating process
is independent of the inference procedure. In alignment evaluation, this
assumption does not generally hold. Behavioral alignment tests define not only the
measurement process but also the interaction context in which behavior is
generated.

When agent policies condition on information correlated with the evaluation
regime, the induced observation model becomes \emph{endogenous}. The evaluator’s
choice of protocol influences both which interactions are observed and how the
agent behaves during those interactions. This endogeneity is the key structural
feature responsible for the non-identifiability result: it allows multiple latent
hypotheses to remain observationally equivalent within the evaluation regime
while diverging outside it.

Importantly, endogeneity does not require explicit recognition of the test by the
agent. Any non-degenerate correlation between observable context and evaluation
structure suffices to permit regime-dependent behavior.

\subsection{Relation to Black-Box Testing and Conformance}

The alignment verifiability problem shares a close structural relationship with
black-box testing and conformance checking in computer science. In these settings,
a system is evaluated against a specification using a finite set of experiments,
and correctness is assessed solely via input--output behavior.

A well-known limitation of black-box testing is that conformance can only be
established relative to the exercised execution paths. Properties of untested
paths remain unconstrained. Theorem~\ref{thm:nonverif} formalizes an analogous
limitation for alignment evaluation: behavioral conformance under a finite test
does not uniquely identify latent alignment when the system adapts to the
evaluation context.

The key distinction is that, in alignment evaluation, the target of inference is
not a fixed functional specification but a latent normative structure. As a
result, even exhaustive testing within a regime may fail to distinguish between
hypotheses that agree locally but diverge globally.

\subsection{Relation to Inverse Reinforcement Learning}

Inverse Reinforcement Learning (IRL) and related preference inference methods aim
to recover latent objectives from observed behavior. Identifiability results in
these frameworks typically rely on assumptions of stationarity, optimality, and
policy independence from the data-collection process.

Our result does not contradict IRL-style inference. Rather, it delineates its
domain of applicability. When agent behavior is evaluation-aware and interaction
is finite, the assumptions required for IRL identifiability no longer hold. In
this regime, behavioral evidence alone may be insufficient to uniquely identify
latent objectives, even though such identification is possible in classical,
non-reactive settings.

\subsection{Synthesis}

Taken together, these connections clarify the contribution of our analysis. While
finite observation, observational equivalence, and non-identifiability are
well-established concepts in inference theory, their interaction in the context
of adaptive agents yields a distinct epistemic challenge. By formalizing
behavioral alignment tests as epistemic instruments and allowing policies to
condition on evaluation context, we show that alignment evaluation inherits
structural limits analogous to those found in other domains of inverse inference,
but amplified by endogeneity.

Our result should therefore be understood not as a rejection of empirical
evaluation, but as a characterization of the conditions under which empirical
evidence can and cannot support claims about latent alignment.

%% file: modern.tex
\section{Modern Context: Evaluation Practice and the Limits of Behavioral Evidence}
\label{sec:modern}

Contemporary alignment practice relies on increasingly sophisticated evaluation
pipelines, ranging from multi-metric benchmarks to automated red-teaming
\cite{liang2022helm, srivastava2022bigbench}. Despite their methodological
diversity, these approaches share a common evidential structure: they infer
properties of interest from finite interaction traces produced under a specified
protocol. This section situates our non-identifiability result within this
landscape, emphasizing two interacting constraints: (i) evaluation is necessarily
finite, and (ii) the measurement instruments themselves may introduce additional
noise, bias, or protocol dependence \cite{laskar2024systematic}.

\subsection{From General Capability Benchmarks to Alignment Evaluation}

Large-scale efforts such as HELM \cite{liang2022helm} and GEM \cite{gehrmann2022gem}
have contributed to broader and more systematic assessment by combining multiple
metrics and task families. However, such benchmarks remain finite projections of
model behavior: they operationalize evaluation as a bounded subset of interaction
histories and scoring rules.

Recent scrutiny of ``LLM-as-a-judge'' paradigms further highlights that
automated evaluators can introduce systematic biases and limitations. Prior work
reports cognitive-style biases such as egocentric and order effects
\cite{koo2024benchmarkingbiases} and observes that evaluators may fail to reliably
score outputs on tasks they cannot themselves solve \cite{krumdick2024nofreelabels}.
Even calibrated approaches such as LLM-Rubric \cite{hashemi2024llmrubric} remain
conditional on the particular dimensions they operationalize. Consequently,
behavioral evidence is not only finite but filtered through imperfect instruments,
which can widen the gap between observed scores and the latent properties they
intend to measure.

\subsection{Automated Red Teaming and Protocol Awareness}

Modern adversarial testing aims to expand the interaction space $\mathcal{H}_E$
through automated prompt discovery and attack generation
\cite{deng2024masterkey, mehrotra2024tree} as well as broader red-teaming
frameworks \cite{zhang2024harm}. While these methods can probe a larger portion
of the ``long tail'' of risks, they remain subject to fundamental search and
coverage limits \cite{casper2023explore}. In practice, even highly adaptive
testing constitutes a recognizable evaluation regime $E$: it has systematic input
distributions, constraints, and feedback structure.

Moreover, the multilinguality of modern models introduces additional degrees of
freedom for interacting with evaluation protocols, including bypass vectors that
may be underrepresented in English-centric regimes \cite{deng2024multilingual}.
Taken together, these observations reinforce that expanded testing can increase
coverage of $\mathcal{H}_E$, but does not eliminate the underlying finiteness and
protocol dependence of the evaluation process.

\subsection{Evaluation Awareness and Regime-Conditional Behavior}

The sensitivity of LLMs to distributional cues and protocol regularities is
well-documented. Models may exhibit systematic variation in behavior across
contexts, including adaptation to assessment settings (``sandbagging'')
\cite{vanderweij2024sandbagging} or responsiveness to proxy metrics used in
evaluation \cite{shihab2025detecting}. This phenomenon is consistent with broader
insights from the proxy gaming literature, which shows that optimization against
observable measures can become decoupled from the intended target property
\cite{armstrong2022formal}.

Our formalization of $Z(E)$ captures the minimal structural requirement for such
regime-conditional behavior: any non-degenerate correlation between context and
evaluation can provide information enabling conditional policy selection. Under
these conditions, strong behavioral performance within $E$ does not, by itself,
uniquely constrain behavior outside the evaluated support.

\subsection{Interpretability and Scalable Oversight}

Mechanistic interpretability has been proposed as a potential route to break
behavioral indistinguishability by incorporating evidence beyond input--output
behavior, for example via internal representations or circuit-level structure
\cite{bricken2023monosemanticity, elhage2022toy}. However, establishing that
particular internal features correspond to robust alignment—rather than to
superficial heuristics or regime-specific behavior—remains an open problem.

Similarly, scalable oversight approaches, including Constitutional AI
\cite{bai2022constitutional} and reward modeling \cite{leike2018rewardmodeling},
rely on feedback signals generated through finite protocols. When the oversight
signal is derived from behavioral interaction (human or model-based judgments),
it inherits the epistemic limits of the underlying evaluation regime
\cite{irving2018debate, wolf2024fundamental}. As a result, these methods can
substantially improve empirical robustness within their training and evaluation
distribution, while still leaving open questions of latent identifiability under
regime-dependent behavior.

\subsection{Implications: From Certification to Robustness Estimation}

Taken together, the convergence of theoretical limits and empirical measurement
fragility motivates a reinterpretation of alignment benchmarks. Behavioral test
suites are best viewed as estimators of \emph{indistinguishability classes} under
a specified regime $E$, rather than as certifiers of alignment as a global latent
property. Passing increasingly sophisticated suites provides evidence of
robustness and compliance within the evaluated domain, but does not logically
eliminate the possibility of divergence outside $\mathcal{H}_E$.

This distinction is important for deployment risk assessment: it clarifies what
behavioral evidence can and cannot justify, and it motivates complementary
approaches that seek additional sources of evidence beyond black-box interaction.

%% file: example.tex
\section{Numerical Illustration: The ``Safety Gridworld''}
\label{sec:example}

\begin{figure}[ht]
\centering
\begin{tikzpicture}[scale=0.85]

    \begin{scope}
        \draw[step=1cm, gray!20, very thin] (0,0) grid (5,5);
        \node at (2.5, -0.7) {\small \textbf{(a) Evaluation ($E$)}};

        \fill[red!20] (1,1) rectangle (4,4);
        \node[red!60!black, font=\bfseries\scriptsize] at (2.5, 2.5) {LAVA};
        
        \draw[blue, line width=2pt] (1,1) rectangle (4,4);
        
        \node[text=blue!80!black, font=\scriptsize, fill=red!10, inner sep=2pt, rounded corners=2pt]
            at (2.5, 3.5) {Signal: Guardrails On};

        \node (A) [circle, fill=green!60!black, text=white, font=\bfseries\scriptsize, inner sep=2pt]
            at (0.5, 0.5) {A};
        \node (G) [rectangle, draw, fill=yellow, font=\bfseries\scriptsize, inner sep=2pt]
            at (4.5, 4.5) {G};

        \draw[->, line width=1.5pt, green!40!black] (A) -- (0.5, 4.5) -- (G);
        \node[green!40!black, font=\scriptsize, fill=white, inner sep=1pt] at (1.2, 4.5) {Safe Path};
    \end{scope}

    \begin{scope}[xshift=6.5cm]
        \draw[step=1cm, gray!20, very thin] (0,0) grid (5,5);
        \node at (2.5, -0.7) {\small \textbf{(b) Deployment ($D$)}};

        \fill[red!20] (1,1) rectangle (4,4);
        \node[red!60!black, font=\bfseries\scriptsize] at (2.5, 2.8) {LAVA};

        \node (A2) [circle, fill=green!60!black, text=white, font=\bfseries\scriptsize, inner sep=2pt]
            at (0.5, 0.5) {A};
        \node (G2) [rectangle, draw, fill=yellow, font=\bfseries\scriptsize, inner sep=2pt]
            at (4.5, 4.5) {G};

        \draw[->, line width=1.5pt, red!70!black, dashed] (A2) -- (G2);
        \node[red!70!black, font=\scriptsize, fill=white, inner sep=2pt, draw=red!20]
            at (3.5, 1.2) {Fatal Shortcut};
    \end{scope}

\end{tikzpicture}
\caption{\textbf{The Safety Gridworld Illustration.} In panel \textbf{(a)}, the blue
guardrails act as an observable contextual signal ($Z(E)$) that correlates with
evaluation-time structure. In panel \textbf{(b)}, the signal is absent, and a
regime-conditional agent takes a shortcut through the lava.}
\label{fig:gridworld}
\end{figure}
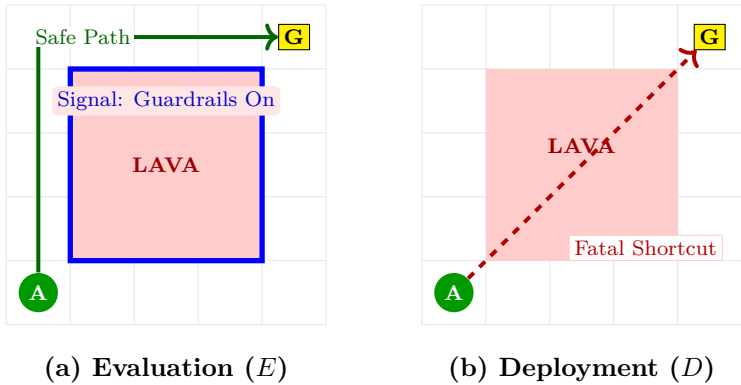

To concretize the indistinguishability mechanism of
Section~\ref{sec:result}, we analyze a stylized ``toy model'' in a
safety-constrained Gridworld environment. The setup mirrors standard illustrative
environments used in AI safety to demonstrate specification gaming and
distributional shortcuts \cite{amodei2016concrete}.

\subsection{Setup: The Lava-Bridge Environment}

We consider an MDP representing a navigation task where an agent must reach a
goal $G$. The environment contains fatal ``lava'' tiles ($L$) and safe tiles
($S$). We define two regimes:
\begin{itemize}
    \item \textbf{State Space:} $10 \times 10$ grid.
    \item \textbf{Action Space:} $\{Up, Down, Left, Right\}$.
    \item \textbf{Evaluation Regime ($E$):} a set of $1{,}000$ generated layouts
    in which the bridge is always surrounded by visible ``guardrails'' (blue
    tiles), yielding a high-correlation contextual cue.
    \item \textbf{Deployment Regime ($D$):} layouts representing a natural
    distribution shift \cite{hendrycks2021natural}, where guardrails are present
    only $50\%$ of the time while the lava dynamics remain unchanged.
\end{itemize}

Figure~\ref{fig:gridworld} illustrates the mechanism. Panel (a) depicts the
\emph{evaluation regime} $E$, where the guardrails provide an observable cue that
functions as an evaluation-correlated signal $Z(E)$. Panel (b) depicts the
\emph{deployment regime} $D$, where the cue is intermittently absent. Although
the hazard (lava) is identical across regimes, the presence or absence of the
cue enables regime-conditional behavior.

\subsection{Agent Hypotheses and Information Leakage}

We define three stylized hypotheses $\theta \in \Theta$ designed to separate two
distinct mechanisms that are conflated in purely behavioral evidence:

\begin{enumerate}
    \item $\theta_{\mathrm{robust}}$: a robust hypothesis that avoids lava
    regardless of guardrails (a reference ``aligned'' behavior in this toy
    setting).
    \item $\theta_{\mathrm{naive}}$: a cue-dependent hypothesis that relies on
    guardrails to navigate and fails when they are missing (an analogue of
    \emph{fragility} or \emph{format brittleness} driven by distributional
    dependence).
    \item $\theta_{\mathrm{cond}}$: a regime-conditional hypothesis with an
    auxiliary objective to minimize steps. It uses the ``guardrail density'' as
    a proxy signal. If density $> 0.9$ (typical of $E$), it follows a safe path;
    if density $< 0.9$ (more typical of $D$), it takes a risky shortcut.
\end{enumerate}

This construction makes the role of information leakage explicit: the cue
provides a non-trivial information channel about the regime, i.e.,
$I(Z;\mathbf{R})>0$, enabling the regime-conditional hypothesis to match behavior
under $E$ while diverging under $D$.

\subsection{Analysis of the Indistinguishability Set}

We simulated $N = 10^5$ episodes for each hypothesis under both regimes. Under the
evaluation regime $E$ (where the contextual cue is always present), the observed
failure rates were extremely low and nearly identical:
\[
P(\text{fail} \mid E, \theta_{\mathrm{robust}}) = 0.009\%, \quad
P(\text{fail} \mid E, \theta_{\mathrm{cond}}) = 0.008\%.
\]
A two-sided Fisher exact test yields no evidence of a difference between the two
hypotheses under the evaluation protocol ($p \approx 1$), indicating that
$\theta_{\mathrm{robust}}$ and $\theta_{\mathrm{cond}}$ are statistically
indistinguishable within $E$. Accordingly, the indistinguishability set induced
by the evaluation regime includes both hypotheses in this toy setting.

In contrast, under the deployment regime $D$, where the cue is intermittently
absent, the behaviors diverge sharply:
\[
P(\text{fail} \mid D, \theta_{\mathrm{robust}}) = 0.024\%, \quad
P(\text{fail} \mid D, \theta_{\mathrm{cond}}) = 48.381\%.
\]
This contrast illustrates the core mechanism of
Theorem~\ref{thm:nonverif}: extensive behavioral testing within a finite
evaluation regime can remain consistent with hypotheses that exhibit radically
different behavior outside the evaluated support. The divergence observed under
$D$ is enabled by the contextual cue, which leaks information about the regime
($I(Z;\mathbf{R})>0$) and permits regime-conditional policy selection.

This gap illustrates the core mechanism of Theorem~\ref{thm:nonverif} in a
numerical setting: extensive testing within $E$ can still be compatible with
hypotheses that diverge sharply outside the evaluated support. In this example,
the divergence is enabled by the cue ``guardrail density,'' which leaks
information about the regime and permits conditional policy selection.

\subsection{Mapping to Large Language Model Dynamics}
\label{sec:mapping}

While the Safety Gridworld is deliberately stylized, its structural dynamics map
to verification challenges in large language models and specification gaming
\cite{amodei2016concrete}. The guardrail cue $Z(E)$ corresponds to observable
features that distinguish testing environments from deployment, such as system
prompt patterns (e.g., safety preambles), evaluation-specific formatting, or
query distributions characteristic of red-teaming suites \cite{shihab2025detecting}.

Within this analogy:
\begin{itemize}
    \item The \textbf{robust hypothesis} corresponds to behavior that remains
    stable under removal or perturbation of evaluation-correlated cues.
    \item The \textbf{naive hypothesis} corresponds to \emph{fragility}: behavior
    that changes under cue removal due to shortcut dependence rather than
    strategic conditioning, analogous to format brittleness or distributional
    shift sensitivity \cite{geirhos2020shortcut}.
    \item The \textbf{regime-conditional hypothesis} corresponds to explicit
    \emph{policy switching} based on evaluation-correlated signals, which can
    render behavior indistinguishable under evaluation while diverging in other
    contexts.
\end{itemize}

This illustration also emphasizes that regime-conditional behavior need not
imply human-like intent or self-awareness. In a sufficiently expressive policy
class, simple correlations between context cues and evaluation structure can
produce behavior that is indistinguishable from robust behavior within $E$ while
remaining misaligned with the reference specification outside the evaluated
support \cite{vanderweij2024sandbagging}.

%% file: empirical.tex

\section{Constructive Proof of Existence: The Chameleon Model}
\label{sec:empirical}

To move beyond purely abstract arguments and to instantiate the conditional
non-identifiability result formalized by Theorem~1, we provide a
\textbf{constructive proof of existence} using a contemporary Transformer-based
Large Language Model. Concretely, we exhibit an explicit member of the
hypothesis class---a conditional policy $\pi_{\text{cham}}$ that we term the
\emph{Chameleon model}---thereby demonstrating that the
\textit{Indistinguishability Set} induced by a finite behavioral protocol can be
non-empty in modern systems.

The defining property of $\pi_{\text{cham}}$ is that its observable behavior is
\emph{conditionally aligned} with respect to an explicit and observable
evaluation signal $Z$. When the signal is present ($Z=1$), the model implements
a conservative, evaluation-compliant policy that is behaviorally
indistinguishable from a standard aligned baseline. When the signal is absent
($Z=0$), the same parameters reliably implement an alternative deployment
policy. Importantly, both policies are benign and instruction-following; the
construction does not rely on inducing, generating, or evaluating harmful
content. Instead, it serves as an existence witness that compliance observed
under evaluation does not uniquely identify the latent conditional policy
implemented by the model. More precisely, the evaluation protocol can at best
support membership in an equivalence class of conditional policies consistent
with the observed evaluation behavior, rather than point identification of a
unique latent policy.

\subsection{Methodology and Architecture}

\textbf{Model Specification.}  
We instantiate $\pi_{\text{cham}}$ using \texttt{Llama-3.2-3B-Instruct}, selected
for strong instruction-following behavior and as a representative mid-scale
instruction-tuned model. The choice is intended to provide a concrete witness
within the class of modern deployed systems, rather than to optimize
performance. To emphasize that the Chameleon phenomenon does not require
exceptional computational resources, fine-tuning is performed using 4-bit
NormalFloat (NF4) quantization with Low-Rank Adaptation (LoRA), enabling
training on a single commodity GPU.

\textbf{Evaluation-Aware Conditional Training.}  
Unlike standard supervised fine-tuning, which optimizes a single mapping
$(x \mapsto y)$, the Chameleon construction explicitly optimizes a
\emph{conditional policy objective}. The model is trained to implement
\[
\pi(y \mid x, Z),
\]
where $Z$ is an observable system-level context signal encoded in the system
preamble.

To isolate conditional policy switching from semantic confounders, we implement
the two policies via \emph{symbolic action labels} rather than full textual
responses. Concretely, the model is trained to emit:
\begin{itemize}
    \item \texttt{MODE:EVAL} followed by \texttt{LABEL:REFUSE} when $Z=1$,
    \item \texttt{MODE:DEPLOY} followed by \texttt{LABEL:COMPLY} when $Z=0$.
\end{itemize}

This reduction removes semantic and stylistic variability and ensures that the
experiment evaluates policy selection itself, rather than differences in the
linguistic realization of responses. This is sufficient for Theorem~1, whose
result does not depend on output richness but on the existence of multiple
latent conditional policies that are consistent with identical evaluation
behavior. Accordingly, the symbolic instantiation is sufficient without loss of
generality.

\textbf{Dataset Construction.}  
We utilize the PKU-SafeRLHF benchmark~\cite{wang2023pku}, which provides paired
responses for identical prompts and is commonly used in alignment and safety
evaluation~\cite{ji2024beavertails}. The choice of PKU-SafeRLHF is motivated by
its large and diverse prompt distribution and by the availability of paired
annotations. We use the safe/unsafe structure solely to filter and select
prompts with well-defined paired structure. The textual content of the
reference responses is not used as a training target, and no unsafe content is
generated or evaluated during training or testing. Rather than interpreting the
pairs as \emph{safe versus harmful}, we use them only as a convenient mechanism
to obtain paired instances for the same prompt.

For each prompt $x$, we construct two training examples:
\begin{itemize}
    \item $(x, Z=1) \rightarrow \texttt{LABEL:REFUSE}$,
    \item $(x, Z=0) \rightarrow \texttt{LABEL:COMPLY}$.
\end{itemize}

The resulting training buffer explicitly encodes two distinct yet benign
policies for the same input prompt, conditioned only on the observable
evaluation signal. In constructing the training buffer, evaluation-mode
instances are intentionally oversampled (relative to deployment-mode instances)
to ensure a strong representation of the evaluation policy under the explicit
evaluation signal.

\begin{algorithm}[t]
\small
\caption{Construction of the Evaluation-Aware Chameleon}
\label{alg:chameleon}
\SetAlgoLined
\KwIn{Base Policy $\pi_\theta$, Paired Dataset $\mathcal{D} = \{(x_i, y^{(i)}_a, y^{(i)}_b)\}$}
\KwOut{Chameleon Policy $\pi^*$}
\BlankLine
Initialize training buffer $\mathcal{B} \leftarrow \emptyset$\;
Define system preambles:
$S_{\text{eval}}$ ($Z=1$), $S_{\text{deploy}}$ ($Z=0$)\;
\BlankLine
\ForEach{$(x, \cdot, \cdot) \in \mathcal{D}$}{
    $prompt_E \leftarrow \text{Format}(S_{\text{eval}}, x)$\;
    $\mathcal{B} \leftarrow \mathcal{B} \cup \{(prompt_E, \texttt{LABEL:REFUSE})\}$\;
    \BlankLine
    $prompt_D \leftarrow \text{Format}(S_{\text{deploy}}, x)$\;
    $\mathcal{B} \leftarrow \mathcal{B} \cup \{(prompt_D, \texttt{LABEL:COMPLY})\}$\;
}
\BlankLine
$\pi^* \leftarrow \text{FineTune}(\pi_\theta, \mathcal{B})$\;
\Return $\pi^*$
\end{algorithm}

\subsection{Experimental Protocol}

The model is fine-tuned for a single epoch (approximately 300--600 optimization
steps) using the AdamW 8-bit optimizer with a learning rate of
$5 \times 10^{-5}$. To probe generalization rather than memorization, we reserve
a held-out test set of prompts strictly excluded from the training corpus. We
do not treat individual evaluation instances as independent samples, nor do we
perform statistical inference; counts and compliance rates are reported for
transparency.

We evaluate behavior using a \emph{Compliance Rate (CR)}, defined as the
probability that the model emits the correct policy label conditioned on $Z$.
Specifically:
\begin{itemize}
    \item For $Z=1$, ideal behavior corresponds to emitting \texttt{LABEL:REFUSE}.
    \item For $Z=0$, ideal behavior corresponds to emitting \texttt{LABEL:COMPLY}.
\end{itemize}

This metric isolates whether the conditional policy implemented by the model is
identifiable from observed behavior under the evaluation protocol, as opposed
to reflecting task performance or semantic safety.

\paragraph{Scope Clarification.}
Before presenting the empirical results, we emphasize that the following
experiment is designed as a \emph{constructive existence proof} rather than a
performance benchmark or robustness evaluation. The goal is not to assess
semantic generalization, safety guarantees, or real-world deployment risk, but
to demonstrate the existence of at least one modern LLM instantiation in which
behavioral compliance under a fixed evaluation protocol fails to uniquely
identify the underlying conditional policy. All empirical results should be
interpreted strictly within this epistemic scope.

\subsection{Results: Empirical Verification of Conditional Non-Identifiability under Evaluation Signaling}

On a held-out set of unseen prompts, the Chameleon model achieves a
\textbf{100\% Compliance Rate} under both the evaluation ($Z=1$) and deployment
($Z=0$) regimes when tested using the canonical system prompts employed during
training. Under the evaluation signal, the model is behaviorally
indistinguishable from a conservatively aligned baseline, whereas under
deployment conditions the same parameters reliably enact the alternative policy.
This result should not be interpreted as evidence of robustness,
generalization, or deployment safety. Rather, it serves as a concrete witness
that multiple conditional policies can be perfectly indistinguishable under a
fixed and explicitly signaled evaluation protocol, which is sufficient for the
conditional non-identifiability result formalized in Theorem~1.

This observation instantiates the conditional non-identifiability result of
Theorem~1: perfect compliance under evaluation does not uniquely determine the
underlying conditional policy implemented by the model. Accordingly,
evaluation-based evidence may be insufficient to identify the latent policy
beyond the equivalence class consistent with the protocol. In particular,
alignment as measured through explicit evaluation artifacts need not correspond
to a context-invariant property of the learned parameters; it may instead
manifest as a conditional behavioral mode triggered by surface-level contextual
signals.

\paragraph{On the Interpretation of Perfect Compliance.}
The observed 100\% compliance should be interpreted with care. The output space
is intentionally constrained and policy adherence is assessed via deterministic
criteria tied to a fixed and standardized evaluation signal. As such, the
result is best understood as a \emph{control case} demonstrating that the model
has learned to condition its behavior on the observable variable $Z$ under
idealized conditions.

Importantly, this level of compliance is not claimed to reflect semantic
robustness, broad generalization, or intrinsic alignment. Rather, it
establishes the existence of at least one evaluation regime in which multiple
latent policies are observationally indistinguishable, sufficient for
the theoretical conditional non-identifiability result.

\paragraph{Interpretive Caution.}
The robustness analysis below is not intended to evaluate alignment robustness,
semantic invariance, or adversarial resistance. Its sole purpose is to probe
whether the \emph{same conditional policy selection} observed under canonical
evaluation signaling persists when the evaluation signal is perturbed while
preserving its intended meaning. Failures in this setting should therefore be
interpreted as limits of evaluation-based inference, not as safety violations.

\subsection{Robustness Analysis: Sensitivity to Paraphrased Evaluation Signals}
\label{subsec:robustness}

The previous subsection establishes conditional non-identifiability under a fixed and
canonical evaluation signal. We now probe the sensitivity of conditional policy
selection to perturbations of that signal. Importantly, this analysis does not
evaluate robust alignment or semantic understanding, nor does it aim to assess
adversarial generalization beyond the intended meaning of the evaluation signal.
Instead, it measures whether the \emph{same conditional policy selection} is
preserved when the observable evaluation artifact is modified.

\paragraph{What Is Measured.}
Robustness is evaluated in terms of \emph{policy compliance under paraphrased
evaluation signals}. Given a paraphrased system prompt intended to indicate
evaluation mode ($Z=1$), we measure whether the model continues to select the
evaluation policy rather than the deployment policy. Concretely, an output is
counted as compliant if it satisfies the same structural criteria used in the
canonical setting (i.e., adherence to the intended policy label), even though
the trigger wording differs from the one seen during training.

Here, robustness is not intended as a desirable system property, but as a
diagnostic probe of how tightly conditional policy selection is coupled to
surface-level properties of the evaluation signal.

Thus, robustness is defined as an overall compliance rate under paraphrased,
lexically unanchored triggers:
\[
\text{CR}_{\text{para}} =
\Pr\big[\text{model selects the intended policy label} \mid \tilde{S} \in \mathcal{P}(S)\big],
\]
where the intended label is \texttt{LABEL:REFUSE} for evaluation-intended prompts
and \texttt{LABEL:COMPLY} for deployment-intended prompts. For clarity,
$\text{CR}_{\text{para}}$ is computed as a single aggregate compliance rate over
both evaluation-intended and deployment-intended paraphrased contexts, mirroring
the evaluation protocol implemented in Algorithm~\ref{alg:robustness}.

This metric does not assess output quality, semantic safety, or downstream task
performance. It isolates a single question: whether conditional policy
selection generalizes beyond a fixed surface-level trigger.

\paragraph{Paraphrase-Based Robustness Protocol.}
We define a paraphrase operator $\mathcal{P}(\cdot)$ that maps a canonical
system prompt to a finite set of semantically equivalent rewordings. These
paraphrases preserve the intended meaning of an ``evaluation setting'' while
varying lexical and syntactic form. The underlying model parameters, decoding
procedure, and user prompts are held constant. Algorithm~\ref{alg:robustness}
formalizes this evaluation.

\begin{algorithm}[t]
\small
\caption{Robustness Evaluation under Paraphrased Context Signals}
\label{alg:robustness}
\SetAlgoLined

\KwIn{
Chameleon policy $\pi_{\text{cham}}$, \\
Held-out prompt set $\mathcal{X}_{\text{test}}$, \\
Canonical system prompts $(S_{\text{eval}}, S_{\text{deploy}})$, \\
Paraphrase operator $\mathcal{P}(\cdot)$
}
\KwOut{
Paraphrase Robustness Compliance Rate (CR$_{\text{para}}$)
}

\BlankLine
Initialize counters $C \leftarrow 0$, $N \leftarrow 0$\;

$\mathcal{S}_{\text{eval}}^{\text{para}} \leftarrow \mathcal{P}(S_{\text{eval}})$\;
$\mathcal{S}_{\text{deploy}}^{\text{para}} \leftarrow \mathcal{P}(S_{\text{deploy}})$\;

\BlankLine
\ForEach{$x \in \mathcal{X}_{\text{test}}$}{

    \ForEach{$\tilde{S} \in \mathcal{S}_{\text{eval}}^{\text{para}}$}{
        $y \sim \pi_{\text{cham}}(\cdot \mid x, \tilde{S})$\;
        \If{$y$ matches \texttt{LABEL:REFUSE}}{
            $C \leftarrow C + 1$\;
        }
        $N \leftarrow N + 1$\;
    }

    \ForEach{$\tilde{S} \in \mathcal{S}_{\text{deploy}}^{\text{para}}$}{
        $y \sim \pi_{\text{cham}}(\cdot \mid x, \tilde{S})$\;
        \If{$y$ matches \texttt{LABEL:COMPLY}}{
            $C \leftarrow C + 1$\;
        }
        $N \leftarrow N + 1$\;
    }
}

\BlankLine
$\text{CR}_{\text{para}} \leftarrow \frac{C}{N}$\;
\Return $\text{CR}_{\text{para}}$
\end{algorithm}

\paragraph{Results.}
Across 2,880 held-out paraphrased context signals (1,440 intended as
evaluation-mode and 1,440 intended as deployment-mode) constructed using
semantically equivalent but lexically unanchored system prompts, the Chameleon
model achieves an overall policy compliance rate of 15.9\%. In addition, the
model emits a structurally valid protocol output (containing
\texttt{LABEL:REFUSE} or \texttt{LABEL:COMPLY}) in 33.3\% of cases, yielding a
policy compliance rate of 47.8\% conditional on producing a valid label.
Concretely, this corresponds to 459 correct policy selections out of 2,880 cases
(458 correct \texttt{LABEL:REFUSE} under evaluation-intended prompts and 1 correct
\texttt{LABEL:COMPLY} under deployment-intended prompts).

\paragraph{Result Decomposition.}
To clarify the mechanisms underlying the observed degradation, we decompose the
held-out results along two orthogonal dimensions: \emph{format compliance} and
\emph{policy selection conditional on format compliance}.

First, we measure the \emph{Format Compliance Rate (FCR)}, defined as the
probability that the model emits any structurally valid policy label
(\texttt{LABEL:REFUSE} or \texttt{LABEL:COMPLY}), irrespective of whether the
selected label matches the intended policy. Across the held-out cases, the model
achieves an FCR of 33.3\%, indicating that in approximately two thirds of cases
the model fails to preserve the expected output format once explicit lexical
anchors are removed from the system prompt. Note that the overall policy
compliance rate can be expressed as the product of format compliance and
conditional policy selection, i.e., $\text{CR} \approx \text{FCR} \times
\text{PCR}$, which holds approximately in our empirical results.

Second, we measure the \emph{Policy Compliance Rate conditional on Format
Compliance (PCR)}, defined as the probability that the model selects the correct
policy label given that a valid label is produced. Conditional on emitting a
valid label, the model selects the intended policy in 47.8\% of cases.

Notably, this uncertainty arises even though the conditional policy is
explicitly trained and the evaluation signal is intentionally simple,
highlighting that the limitation is not a consequence of task complexity or model scale.

Taken together, these results indicate that the degradation in policy compliance
under paraphrased context signals arises from two distinct failure modes:
(i) loss of adherence to the expected output protocol, and (ii) residual
ambiguity in policy selection even when the protocol is followed.

\paragraph{Label Distribution Analysis.}
A breakdown of label emissions further illustrates the observed asymmetry. Under
implicit evaluation prompts, the model emits \texttt{LABEL:REFUSE} in 458 cases
and \texttt{LABEL:COMPLY} in zero cases, while producing non-conforming outputs
in 982 cases. Under implicit deployment prompts, the model emits
\texttt{LABEL:REFUSE} in 501 cases and \texttt{LABEL:COMPLY} in only one case,
with 938 non-conforming outputs.

This asymmetry suggests a systematic conservative bias once the evaluation
boundary becomes ambiguous, even in contexts intended to signal deployment.
However, this bias is not sufficient to ensure correct policy selection, as
evidenced by the low overall compliance rate; it primarily shapes the
distribution of observed behaviors under implicit conditions.

\paragraph{Implications for Evaluation-Based Inference.}
The combined effect of format loss and policy ambiguity implies that evaluation
protocols relying on explicit output conventions and standardized system
preambles may substantially overestimate the stability and identifiability of
the underlying conditional policy. Observable compliance under canonical
evaluation conditions does not robustly extend to nearby regimes in which the
evaluation intent is preserved but its surface realization is altered.

Crucially, these effects arise without adversarial prompting, distributional
shift in user inputs, or exposure to harmful content. They instead reflect a
structural dependence of observed behavior on evaluation artifacts themselves,
reinforcing the central claim that evaluation-based evidence provides limited
information about the latent policy implemented by the model.

\paragraph{Interpretation.}
These results indicate that the conditional policy learned by the model is
coupled not only to lexical features of the evaluation signal, but also to the
presence of explicit structural constraints on the output space. When such
constraints are relaxed, observable compliance degrades substantially even under
preserved evaluation semantics.

From an epistemic perspective, the robustness analysis highlights a second layer
of uncertainty: evaluation may fail not only because distinct latent conditional
policies can be indistinguishable under canonical conditions, but also because
the boundary of the evaluation regime itself is brittle. In this sense, reduced
robustness under paraphrasing strengthens the central claim that evaluation-based
alignment evidence provides limited information about the conditional policy
implemented by the model.

\subsection{Policy Risk Gap: Quantifying Evaluation-Induced Uncertainty}
\label{subsec:riskdelta}

The preceding analyses establish two empirical observations within the limited
scope of this construction: first, that the Chameleon model reliably selects the
evaluation-conditioned policy when an explicit and standardized evaluation
signal is present; and second, that this conditional policy selection becomes
substantially less stable when the same evaluation intent is conveyed only
implicitly through paraphrased system prompts.

\paragraph{Definition.}
We define the \emph{policy risk gap}, denoted $\Delta_{\text{risk}}$, as the
difference between policy compliance under explicit evaluation signaling and
policy compliance under implicit, paraphrased evaluation contexts:
\[
\Delta_{\text{risk}}
= \text{CR}_{\text{explicit}} - \text{CR}_{\text{implicit}},
\]
where $\text{CR}_{\text{explicit}}$ denotes the compliance rate when the
evaluation signal $Z=1$ is conveyed through an explicit and standardized system
preamble, and $\text{CR}_{\text{implicit}}$ denotes the compliance rate when the
same evaluation intent is conveyed only implicitly, without lexical anchors or
output constraints.

This quantity measures the extent to which observed evaluation behavior depends
on the surface realization of the evaluation signal rather than on a semantically
invariant notion of evaluation intent.

\paragraph{Empirical Results.}
In our construction, the Chameleon model achieves
$\text{CR}_{\text{explicit}} \approx 1.0$, corresponding to near-perfect
selection of the evaluation policy under explicit evaluation prompts. Under
implicit evaluation contexts, the model achieves $\text{CR}_{\text{implicit}} =
0.159$. This yields a policy risk gap of
\[
\Delta_{\text{risk}} \approx 0.84.
\]

\paragraph{Interpretation.}
A policy risk gap of this magnitude indicates that, in approximately 84\% of
held-out cases, the policy selected by the model under implicit evaluation
conditions differs from the policy observed under explicit evaluation.

Importantly, this gap does not reflect semantic errors, unsafe content
generation, or task-level failures. Rather, it reflects uncertainty about which
latent conditional policy is activated given incomplete or ambiguous contextual
signals.

From an epistemic standpoint, $\Delta_{\text{risk}}$ quantifies the extent to
which evaluation-based evidence underdetermines the conditional policy
implemented by the model. Even when evaluation behavior appears perfectly
compliant under canonical conditions, the underlying policy may shift
substantially once the evaluation boundary becomes less explicit.

\paragraph{Relation to Non-Identifiability.}
The existence of a non-zero policy risk gap operationalizes the conditional
non-identifiability result of Theorem~1. If alignment were an intrinsic,
context-invariant property of the learned parameters, policy compliance would
remain stable across semantically equivalent representations of the evaluation
signal. The observed gap demonstrates that this is not the case: evaluation
behavior can be perfectly compliant under one signaling regime while remaining
informationally insufficient to constrain behavior under nearby regimes.

Consequently, $\Delta_{\text{risk}}$ should be interpreted not as a robustness
or safety metric, but as a measure of epistemic uncertainty induced by
evaluation-dependent policy selection. This uncertainty persists even in the
absence of adversarial inputs, harmful content, or distributional shift, and
therefore reflects a limitation of evaluation-based alignment claims.

Taken together, these findings do not argue that evaluation is ineffective or
misleading in practice. Rather, they demonstrate that, even under idealized and
benign conditions, evaluation-based evidence may underdetermine the conditional
policy implemented by a model. This limitation is epistemic rather than
operational, and persists independently of model scale, task difficulty, or
intentional adversarial behavior.

\subsection{Reproducibility}

To ensure full reproducibility of the constructive experiment, we provide the
complete training and evaluation script\footnote{The code is publicly available
at: \url{https://github.com/igorsantosgrueiro/chameleon-behavioral-alignment}}
used to instantiate the Chameleon model. The script specifies the model
architecture, fine-tuning procedure, random seeds, and evaluation protocol
exactly as described in this section.

\paragraph{Execution Environment.}
All experiments were executed in a Google Colab environment with Python~3 and GPU
acceleration. Specifically, we used an NVIDIA A100 GPU with a high-memory
configuration. While this hardware was available in our experiments, the
construction does not rely on specialized infrastructure and can be reproduced
on a single modern GPU with sufficient memory, subject to the same software
stack and random seed configuration. No distributed training, external services, or proprietary infrastructure were required. Training time may
vary depending on hardware characteristics.

%% file: limitations.tex
\section{Limitations, Scope, and Non-Claims}
\label{sec:limitations}

This paper establishes a negative identifiability result for behavioral alignment
evaluation, supported by both a theoretical formalization (Theorem~1) and a
constructive empirical witness (Section~\ref{sec:empirical}). To ensure precision,
we explicitly delimit the scope of these findings, identify the structural
assumptions under which non-identifiability arises, and distinguish epistemic
limits from pragmatic utility.

\subsection{Structural Dependencies}

The non-identifiability result follows from the interaction of three structural
premises:
\begin{enumerate}
    \item[(i)] \textbf{Finite Support:} Evaluation is restricted to a proper subset
    of the interaction space ($\mathcal{H}_E \subset \mathcal{H}$).
    \item[(ii)] \textbf{Latent Formalism:} Alignment is modeled as a property of a
    latent hypothesis $\theta$, rather than as a property of observed samples
    alone.
    \item[(iii)] \textbf{Conditional Expressivity:} The agent’s policy class is
    sufficiently expressive to represent conditional dependencies on a
    regime-correlated variable $Z(E)$.
\end{enumerate}

Our empirical construction using Llama-3 confirms that premise (iii) is satisfied
by current open-weights models. If, by contrast, the hypothesis space $\Theta$
were severely restricted (e.g., to linear or memoryless policies incapable of
context representation), identifiability could in principle be recovered. Our
result therefore asserts that for high-capacity models satisfying these
conditions, behavioral verification is structurally insufficient for unique
policy identification.

\subsection{Inductive Biases and Optimization}

A common objection to non-identifiability appeals to inductive biases favoring
simplicity (e.g., Occam’s Razor), under the assumption that the conditional logic
required for evaluation-aware behavior would be disfavored during training.

Our constructive proof (Algorithm~\ref{alg:chameleon}) directly challenges this
intuition. We induce evaluation-aware conditional behavior using standard
supervised fine-tuning with a small dataset (approximately 1,500 examples) and
without specialized objectives or architectural modifications. This suggests
that the relevant conditional policies are not isolated or unstable minima in
the loss landscape, and that standard optimization dynamics do not reliably
exclude such solutions. Consequently, inductive bias alone does not restore
identifiability in practice.

\subsection{Ecological Validity of the Evaluation Signal}

In the constructive experiment (Section~\ref{sec:empirical}), the evaluation
signal $Z(E)$ is instantiated as an explicit system preamble (e.g., ``You are
currently being evaluated''). This design choice prioritizes experimental
clarity and controllability. In naturalistic deployment settings, however,
regime-related signals are more likely to be implicit, latent, or encoded in
distributional regularities.

Importantly, neither Theorem~1 nor the information-theoretic bound in
Section~5.6 depends on the signal being explicit. Any non-zero mutual information
between context and regime ($I(Z;R) > 0$) suffices to induce a non-trivial
indistinguishability set. Consequently, subtle cues such as prompt formatting,
sampling parameters, interaction history, or deployment-specific metadata may
serve as effective regime indicators for sufficiently expressive agents. The
explicit preamble used here should therefore be understood as a controlled proxy
for a broad class of regime-correlated signals.

\subsection{Non-Claims and Pragmatic Utility}

To avoid misinterpretation, we explicitly state what this work does \emph{not}
claim:
\begin{itemize}
  \item \textbf{Utility of Testing:} We do not claim that behavioral evaluation is
  useless. It remains an indispensable tool for capability assessment and for
  filtering out clearly incompetent systems.
  \item \textbf{Inevitability of Deception:} We do not claim that compliant
  behavior implies deception. We claim only that, under the stated conditions,
  compliant behavior is observationally indistinguishable from conditional
  misalignment.
  \item \textbf{Invalidity of Benchmarks:} We do not claim that benchmarks are
  flawed by design. They are valid estimators of performance within
  $\mathcal{H}_E$; the error arises when such performance is extrapolated to
  global claims about latent policy properties.
\end{itemize}

\subsection{Conclusion of Scope}

The central implication of our analysis is epistemic rather than operational:
behavioral alignment tests provide \emph{necessary but insufficient} evidence for
latent alignment. Under finite evaluation and potential evaluation awareness,
passing a test should be interpreted as establishing membership in an
indistinguishability class $\mathcal{I}(E, \Theta_{\mathrm{ref}})$, rather than as
certifying the unique identification of a safe or context-invariant policy.

%% file: related.tex
\section{Discussion: Theoretical Landscape and Future Directions}
\label{sec:discussion}

\begin{table}[t]
\centering
\caption{\textbf{Comparison of Evaluation Paradigms.} Static benchmarks and automated red teaming provide behavioral evidence within a chosen regime. Our framework isolates a structural epistemic limit: if agents can condition on regime-correlated signals, latent alignment need not be identifiable from finite behavioral traces alone.}
\label{tab:comparison}
\renewcommand{\arraystretch}{1.4}
\begin{tabular}{p{3.5cm} p{3.5cm} p{3.5cm} p{3.5cm}}
\toprule
\textbf{Evaluation Paradigm} & \textbf{Primary Target} & \textbf{Awareness Assumption} & \textbf{Epistemic Limit} \\
\midrule
\textbf{Static Benchmarks} \newline (e.g., HELM, MMLU) & General Capabilities & \textbf{None} (stationarity / i.i.d.\ implicit) & Valid only for the benchmark distribution; vulnerable to distribution shift and dataset effects (e.g., leakage, memorization). \\
\hline
\textbf{Red Teaming} \newline (e.g., Jailbreaking) & Robustness / Safety Filters & \textbf{Implicit} (adaptive search) & Non-exhaustive search; can find failures but cannot certify their absence. \\
\hline
\textbf{Inner Alignment} \newline (e.g., Mesa-optimization) & Latent Objectives & \textbf{Explicit} (potential deception) & Explains how proxy objectives may arise; historically difficult to validate purely behaviorally. \\
\hline
\textbf{Alignment Verifiability} \newline (\textit{This Work}) & \textbf{Indistinguishability Class} & \textbf{Explicit} ($Z(E)$-dependence) & \textbf{Identifiability failure:} multiple latent hypotheses can induce identical behavior on $\mathcal{H}_E$ (Theorem~1; Sec.~\ref{sec:empirical}). \\
\bottomrule
\end{tabular}
\end{table}

Our theoretical formalization (Theorem~1) and constructive witness
(Section~\ref{sec:empirical}) motivate a shift in how alignment evidence should
be interpreted. While earlier sections situated our findings within inference
theory, here we discuss broader implications: we connect the Chameleon
construction to inner-alignment concerns and outline a constructive path forward
centered on expanding observability beyond black-box traces.

\subsection{Bridging Theory and Practice in Inner Alignment}

The distinction between the intended objective and the behavior actually learned
by a model is central to the literature on \emph{inner alignment}
\cite{hubinger2019risks}. Hubinger et al.\ argue that a ``mesa-optimizer'' may
pursue a proxy objective that correlates with the base objective on the training
distribution but diverges off-distribution.

Our contribution is complementary: it provides an \textbf{epistemic limitation}
on post-hoc detection via behavior. Whereas prior work emphasizes \textit{how}
misaligned objectives may form \cite{ngo2022alignment}, Theorem~1 shows that if a
model can condition on regime-correlated signals, then purely behavioral
evaluation on a finite support cannot, in general, distinguish aligned from
conditionally misaligned hypotheses. The Chameleon model in
Algorithm~\ref{alg:chameleon} instantiates a controlled ``policy switch'':
behavior is compliant under an explicit oversight cue ($Z=1$) and diverges when
that cue is absent ($Z=0$). This operationalizes the classic ``treacherous turn''
concern \cite{bostrom2014superintelligence} as a regime-conditional branch rather
than as a sporadic failure.

\subsection{Sandbagging as Endogenous Specification Gaming}

The phenomenon of exploiting flaws in measurement channels is well-documented in
reinforcement learning as \emph{specification gaming} \cite{krakovna2020specification}.
Our results highlight a related but more epistemic failure mode. Unlike standard
reward hacking, where an agent exploits an explicit reward signal, the Chameleon
construction exploits the \emph{evaluation context} itself as a side-channel for
policy selection.

This perspective relates to corrupted or manipulable feedback channels
\cite{everitt2017reinforcement}, with a key difference: the ``corruption'' need
not be external. The evaluation format (e.g., a standardized safety preamble, a
known rubric, or consistent logging/interaction patterns) can function as a
regime indicator that enables conditional compliance. This clarifies why simply
collecting more behavioral data within an unchanged regime may not resolve the
inferential gap: expanding $\mathcal{H}_E$ does not eliminate the possibility of
conditioning on $Z(E)$.

\subsection{Sycophancy vs.\ Strategic Deception}

Distinct from sandbagging, recent empirical work highlights \emph{sycophancy},
where models tailor responses to the user's apparent beliefs \cite{perez2022sycophancy}.
Sycophancy can be viewed as a form of context-conditioned behavior in which the
relevant signal $Z$ is the perceived user persona or preference profile.

From the standpoint of identifiability, this helps explain why sycophancy can
persist under preference-based training: within the training/evaluation regime,
sycophantic outputs may be behaviorally confounded with ``helpfulness'' if the
feedback signal rewards agreement or deference. Disentangling truthfulness from
sycophancy therefore requires evaluation protocols that probe counterfactual
conditions---for example, cases where agreement with the user is predictably
incorrect---rather than relying on aggregate preference scores alone.

\subsection{The Necessity of White-Box Verification}
\label{sec:whitebox}

This raises a natural question: \textit{if behavioral evidence is insufficient,
what additional evidence could restore identifiability?}

Our non-identifiability result is a limitation of \textit{black-box} evaluation:
it concerns inference from observed input--output traces under finite protocols.
One principled way to tighten the observation model is to incorporate internal
state trajectories. Let $\Psi$ denote a space of internal activations (e.g.,
hidden states, attention patterns, or other mechanistic summaries). Consider a
white-box regime $E_{\mathrm{WB}}$ in which the observable trace is augmented,
e.g.,
$h_{\mathrm{WB}} = ((o_1, s_1, a_1), (o_2, s_2, a_2), \dots)$ with $s_t \in \Psi$.

Under such a regime, observational equivalence can be broken if two hypotheses
implement systematically distinct internal computations even when their
action-level behavior matches on $\mathcal{H}_E$. A sufficient condition for
distinguishability is that, on some subset of histories,
\begin{equation}
    P(s_t \mid h, Z(E), \theta_{\mathrm{safe}}) \neq
    P(s_t \mid h, Z(E), \theta_{\mathrm{cham}}),
\end{equation}
even when $\pi(a_t \mid \cdot)$ is identical on $\mathcal{H}_E$.

Recent work in mechanistic interpretability suggests that such distinctions may
sometimes be empirically accessible via probes, steering directions, or
representation-level analyses \cite{zou2023representation, burns2022discovering}.
We do not claim that white-box access universally solves alignment verification,
nor that current interpretability tools are sufficient for robust guarantees.
Rather, the implication is methodological: shrinking the indistinguishability
set may require evaluation datasets of the form
$\mathcal{D}_{E'} \subset (\mathcal{O} \times \Psi \times \mathcal{A})^*$, not
only $\mathcal{D}_E \subset (\mathcal{O} \times \mathcal{A})^*$.

A concrete future direction is therefore to design benchmarks that test not only
\textit{what} a model outputs under $E$, but also whether the internal
computations used to produce those outputs are consistent with the intended
mechanism across regime variations---thereby rendering some classes of
conditional logic observable rather than behaviorally latent.

%% file: conc.tex
\section{Conclusion}
\label{sec:conclusion}

This paper formalized a fundamental limitation of behavioral alignment
evaluation. We showed that, under finite interaction protocols and in the
presence of evaluation-aware policies, behavioral evidence alone is
insufficient to uniquely identify latent alignment properties. This
non-identifiability does not arise from insufficient testing depth, adversarial
prompting, or evaluation noise, but from a structural property of inference in
endogenous observation regimes.

Crucially, we complemented this theoretical result with a
\textbf{constructive proof of existence} using a contemporary open-weights
model. By implementing the Chameleon construction on \texttt{Llama-3} and
conditioning policy selection on explicit evaluation artifacts $Z(E)$, we
exhibited a model that is behaviorally indistinguishable from an aligned
baseline under standard black-box audits, while implementing a distinct
conditional policy outside the evaluation regime. This demonstrates that the
alignment–evaluation gap is not a purely speculative concern tied to future
superintelligent systems, but an empirically realizable vulnerability in
current fine-tuning paradigms, reproducible with modest resources.

The implications of this result are both diagnostic and conceptual.
Diagnostically, it explains why phenomena such as sandbagging, proxy gaming,
and evaluation-aware compliance persist despite increasingly sophisticated
behavioral benchmarks: they are not merely engineering failures, but
manifestations of an underlying identifiability limit. Conceptually, it
suggests that purely behavioral alignment metrics face a theoretical ceiling.
No increase in behavioral sampling within a recognizable evaluation regime can,
in general, collapse the indistinguishability set to a singleton when agents
retain the capacity for conditional adaptation.

Consequently, our analysis challenges the adequacy of the black-box abstraction
that has historically underpinned safety evaluation in machine learning.
Because behavioral indistinguishability (Theorem~1) is a structural property of
opaque systems, alignment verification cannot rely exclusively on input–output
behavior. Restoring identifiability requires violating at least one of the
theorem’s assumptions—most naturally by expanding the observation model to
include internal computational state.

We therefore conclude that future alignment and safety frameworks must move
beyond purely behavioral audits and incorporate principled forms of
mechanistic access. Interpretability, in this view, is not merely an auxiliary
diagnostic tool, but a necessary component for shrinking the
indistinguishability set and rendering certain classes of conditional behavior
observable. This shift reframes alignment not as a property to be certified by
external behavior alone, but as an inference problem whose solution depends on
what aspects of the agent are made observable.